\documentclass{article}

 \usepackage[final]{neurips_2019}

\usepackage[utf8]{inputenc} 
\usepackage[T1]{fontenc}    
\usepackage{hyperref}       
\usepackage{url}            
\usepackage{booktabs}       
\usepackage{amsfonts}       
\usepackage{nicefrac}       
\usepackage{microtype}      
\usepackage{graphicx}
\usepackage{natbib}
\setcitestyle{square,numbers}
\title{UATTA-EB: Uncertainty-Aware Test-Time Augmented Ensemble of BERTs for Classifying Common Mental Illnesses on Social Media Posts}

\author{
  Pratinav Seth \thanks{Manipal Institute of Technology, Manipal Academy of Higher Education, Manipal, India}\\
  Dept. of Data Science and Computer Applications\\
  \texttt{seth.pratinav@gmail.com} 
   \And
   Mihir Agarwal \footnotemark[1] \\
   Dept. of Data Science and Computer Applications\\
   \texttt{mihiragarwal1423@gmail.com} \\
}
\begin{document}

\maketitle

\begin{abstract}
  Given the current state of the world, because of existing situations around the world, millions of people suffering from mental illnesses feel isolated and unable to receive help in person. Psychological studies have shown that our state of mind can manifest itself in the linguistic features we use to communicate. People have increasingly turned to online platforms to express themselves and seek help with their conditions. Deep learning methods have been commonly used to identify and analyze mental health conditions from various sources of information, including social media. Still, they face challenges, including a lack of reliability and overconfidence in predictions resulting in the poor calibration of the models. To solve these issues, We propose UATTA-EB: Uncertainty-Aware Test-Time Augmented Ensembling of BERTs for producing reliable and well-calibrated predictions to classify six possible types of mental illnesses- None, Depression, Anxiety, Bipolar Disorder, ADHD, and PTSD by analyzing unstructured user data on Reddit.

\end{abstract}
\section{Introduction}
Mental illness is a type of health condition that alters a person's intellect, feelings, or behavior (or all three) and has been shown to affect an individual's physical health. Depression, schizophrenia, attention deficit hyperactivity disorder (ADHD), autism spectrum disorder (ASD), and other mental health issues are widespread today, with an estimated 450 million worldwide suffering from such problems\cite{Sayers2001TheWH}. Unlike other chronic conditions, which are diagnosed using research facility tests and measurements, mental illnesses are regularly diagnosed using an individual's self-report surveys designed to detect specific patterns of feelings or social interactions\cite{Hamilton1967DevelopmentOA}. Many people have expressed clinical anxiety or depression during these uncertain times when COVID-19 plagues the world. This could be due to the lockdown, limited social activities, higher unemployment, economic depression, and work-related fatigue\cite{Ameer2022MentalIC}. According to the American Foundation for Suicide Prevention, people are more likely to experience anxiety (53\%) and sadness (51\%) than before covid-19 was widely available\cite{Ameer2022MentalIC}. Social media has altered social interaction over the last decade. Individuals effectively communicate their day-to-day activities, experiences, hopes, emotions, and so on, sharing data and news and generating massive amounts of data online\cite{Murarka2021ClassificationOM}. 
These texts offer information that can be used to identify the mental health of individuals. Early detection of mental health problems is critical in better understanding and treating mental health conditions. However, many deep learning-based approaches have been introduced recently for the early detection of these conditions. Still, most of them suffer from overconfidence and are poorly calibrated, impacting the model's reliability. 

To solve this problem, we introduce a reliable end-to-end architecture: UATTA-EB: Uncertainty-Aware Test-Time Augmented Ensemble of BERTs for Classifying Common Mental Illnesses on Social Media Posts for producing reliable and well-calibrated predictions.
\begin{figure*}[t]
  \centering
\includegraphics[width=0.75\textwidth]{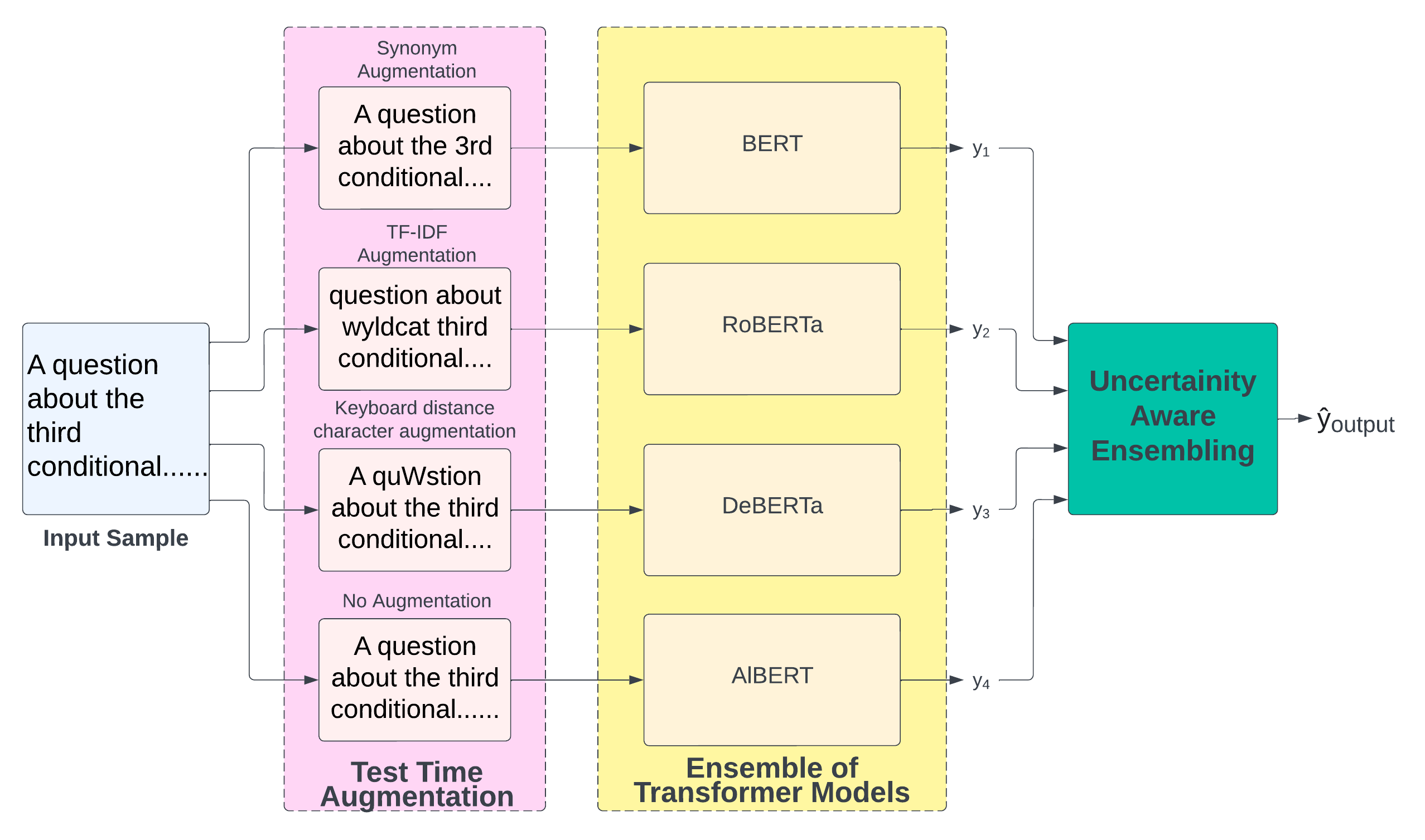}
   \caption{UATTA-EB : Model Architecture}
   \label{fig:modeldesign}
\end{figure*}

\section{Previous Works}
People have recently started communicating and seeking assistance with mental health difficulties on social media. This has inspired researchers to use the data and a range of NLP and ML techniques to assist those who might need assistance. Several methods for studying mental health have been developed through monitoring user behavior on social media. Initially, many researchers concentrated on data from Facebook, Twitter\cite{Benton2017MultiTaskLF}\cite{coppersmith-etal-2015-clpsych}\cite{husseini-orabi-etal-2018-deep}, Instagram, Flickr, and other social media platforms to research mental health. Recently, the attention has turned to the Reddit\cite{Gkotsis2017CharacterisationOM}\cite{Kim2020ADL}\cite{Zirikly2019CLPsych2S}.
Various approaches to mental health text analysis have been used, from traditional ML to advanced DP, including \cite{Benton2017MultiTaskLF}\cite{coppersmith-etal-2015-clpsych}\cite{Amjad2021ThreateningLD}\cite{Amjad2020BendTT}. Much work has been centered around applying Natural Language Processing to detect and classify mental illnesses. 

Data Augmentation is a common practice for artificially expanding a given dataset, in which new data are generated by transforming existing examples \cite{Lu2022ImprovedTC}. While Data Augmentation is frequently performed during model training, recent work has shown that using Data Augmentation at inference, or Test-Time Augmentation (TTA)\cite{Shanmugam2021BetterAI}, can improve Model Accuracy \cite{Krizhevsky2012ImageNetCW}\cite{Szegedy2015GoingDW}\cite{Matsunaga2017ImageCO}, Robustness \cite{Prakash2018DeflectingAA}\cite{Song2018PixelDefendLG} \cite{Cohen2019CertifiedAR}, and Uncertainty Estimates \cite{Matsunaga2017ImageCO}\cite{Smith2018UnderstandingMO}\cite{Wang2019AleatoricUE}. Text Augmentation at test time has been applied to several tasks, including Abstractive Summarization \cite{Parida2019AbstractTS}\cite{Zhu2022TransformingWI}, Question Answering \cite{Longpre2019AnEO}\cite{Yang2019DataAF}\cite{Singh2019XLDACD}, Sequence Tagging \cite{Ding2020DAGADA}\cite{Sahin2018DataAV}\cite{Dai2020AnAO}, Parsing \cite{Jia2016DataRF}\cite{Yu2021GraPPaGP}\cite{Andreas2020GoodEnoughCD}, Grammatical Error Correction \cite{Boyd2018UsingWE}\cite{Zhang2019SequencetosequencePW}\cite{Wang2022ControllableDS}, and Neural Machine Translation \cite{Sennrich2016ImprovingNM}\cite{Xia2019GeneralizedDA}.

Studies show that traditional neural networks are prone to over-confidence \cite{Guo2017OnCO}\cite{Ovadia2019CanYT}. In classification tasks, a poorly calibrated network can place a high probability on one of the classes, even when the predicted class is incorrect \cite{Ayhan2020ExpertvalidatedEO}. The uncertainty estimation issue is especially important to trust confident model predictions for screening automation and referring uncertain cases for manual intervention of a medical expert. Even though fine-tuned pre-trained transformers \cite{Devlin2019BERTPO} have achieved state-of-the-art accuracy on text classification tasks, they still suffer from the same over-confidence problem of traditional neural networks, making the prediction untruthful \cite{Hendrycks2020PretrainedTI}. Overall, there is less work in quantifying uncertainty in the NLP domain. Among them, there are Bayesian and non-Bayesian methods. Bayesian probability theory offers a sound mathematical framework to design machine learning models with an inherent and explicit notion of uncertainty \cite{Yang2021UncertaintyQA}. Instead of resulting in a single per-class probability, such models can estimate the moments of the output distribution for every class, including mean and variance. Multiple probabilistic and Bayesian methods, \cite{Graves2011PracticalVI}\cite{pmlr-v37-blundell15}\cite{HernndezLobato2015ProbabilisticBF}\cite{Blum2015VariationalDA}\cite{Gal2016DropoutAA}\cite{Lee2018DeepNN}\cite{Wu2019DeterministicVI}\cite{Pearce2020UncertaintyIN} and non-Bayesian such as \cite{Osband2016RiskVU}\cite{Sarawgi2020WhyHA}\cite{Lakshminarayanan2017SimpleAS}\cite{Kendall_2018_CVPR}\cite{chang2017active} and \cite{Dusenberry2020AnalyzingTR} have been proposed to quantify the uncertainty estimates. The ensemble of networks can further improve the performance of models.

\section{Methodology}
\begin{table*}[t]
\centering
\begin{tabular}{llllll}
\hline
Model Architecture                                      & Accuracy & F1 Score & ECE   & MCE   & Brier Score \\ \hline
Baseline (BERT)                                                   & 0.84              & 0.84              & 0.046          & 0.259          & 0.29                 \\
Baseline (RoBERTa)                                                & \textbf{0.86}              & \textbf{0.86}              & 0.039          & 0.398          & 0.25                 \\
Baseline (DeBERTa)                                                & 0.82              & 0.82              & 0.050          & 0.332          & 0.34                 \\
Baseline (ALBERT)                                                 & 0.83              & 0.83              & 0.047          & 0.272          & 0.31                 \\
UA-ENS(Uncertainty Aware Ensemble)                                & 0.85              & \textbf{0.86}             & 0.023          & 0.189          & \textbf{0.23}        \\
UATTA-EB(Proposed) & 0.85              & \textbf{0.86}             & \textbf{0.018} & \textbf{0.122} & \textbf{0.23}        \\ \hline
\end{tabular}
\caption{Evaluation Metrics and Uncertainty Metrics for various model architectures on the test set. }
  \label{tab:comb}
\end{table*}
\label{section:meth}
Given a training set $(x_n, y_n)_{n=1}^N$ consisting of N i.i.d examples, we define an end-to-end model architecture as depicted in Figure \ref{fig:modeldesign}. The architecture consists of  multiple stages. The text input is augmented in the first stage using the TTAug Section \ref{subsection:tta}. We use the augmented text as input for the Transformer models \cite{wolf-etal-2020-transformers}, which produce individual predictions, which are then ensembled using Uncertainty-Aware Ensemble Section \ref{subsection:uaens} by weighting the predictions based on their uncertainty to get the final output $\hat{y}$.
\subsection{Test Time Augmentation}
\label{subsection:tta}
Test-Time Augmentation \cite{Lu2022ImprovedTC}\cite{Wang2019AleatoricUE} is the aggregation of predictions across transformed examples of test inputs and is an established technique to improve performance and produce well-calibrated classification models \cite{Lu2022ImprovedTC}. It can deliver consistent improvements over current state-of-the-art approaches. The train and test samples are randomly augmented, which accounts for the noise that may arise in real-world data, thus making the model more robust. The transformations used in our approach using \cite{ma2019nlpaug} include Synonym augmentation for $30\%$ of the sample words, TF-IDF word augmentation for $5\%$ of the sample words, and keyboard augmentation of individual characters at $5\%$ of the sample.
\subsection{Uncertainity Aware Ensemble}
\label{subsection:uaens}
We calculate the uncertainty estimation of the output of each model to create an ensemble architecture that is aware of the uncertainty of the predictions while ensembling. The output prediction $\hat{y_i}$ where $i\in(1, N)$ is weighted with its uncertainty. For the final prediction, $\hat{y}$, the weighted ensemble quantifies an uncertainty weighted average \cite{Jaskari2022UncertaintyAwareDL}.
\begin{equation}
\hat{y}\left(\mathbf{x}_n\right)=\frac{\sum_{j=1}^k \frac{1}{\sigma_{h^j}\left(\mathbf{x}_n\right)} \hat{y}_{h^j}\left(\mathbf{x}_n\right)}{\sum_{j=1}^k \frac{1}{\sigma_{h^j}\left(\mathbf{x}_n\right)}}
\end{equation}
Here, $x_{n}$ denotes a nth input sample, $\hat{y}_{h^j}(x_{n})$ is the output of nth input sample for the $j_{th}$ model prediction,$\sigma_{h^j}$ is the estimated uncertainty  corresponding to predictions from the $j_{th}$ model prediction. The uncertainty weights are formulated by taking an inverse of the independent uncertainty metric. \cite{sarawgi2021uncertainty}. Hence, $\hat{y}\left(\mathbf{x}_n\right)$ outputs the final prediction corresponding to $n^{th}$ data point. Here the uncertainty associated with each prediction is quantified using a modified version of LLFU\cite{Lakara2021EvaluatingPU}.
\begin{equation}
\label{sec:llfu}
    \sigma_{h^j} = max(0,\frac{1}{2}log(2\pi\sigma^2(x_{n}))) 
    + 
    \frac{(y_j(x_{n})-\mu(x_{n}))^2}{2\sigma^2(x_{n})}
\end{equation}
where $y_j(x_{n})$ - denotes the prediction corresponding to $jth$ model, $\mu(x_{n})$ refers to the mode of predictions from all the models in ensemble and $\sigma^2(x_{n})$ refers to standard deviation of predictions of models in ensemble for the $n^{th}$ data point. 
\section{Experiments}
We conduct a series of experiments to study whether the architecture described in Section \ref{section:meth} can correctly classify mental illnesses. We have used four pretrained Language model backbones each followed by a fully connected layer and Softmax from HuggingFace \cite{wolf-etal-2020-transformers} in the Ensemble as described in Section \ref{subsection:sde} along with their respective tokenizers. The maximum sequence length is 512. We used CrossEntropy loss with the Adam \cite{Kingma2015AdamAM} optimizer with a $10^{-5}$ learning rate and batch size of 16.
\subsection{Dataset}
We used \cite{Murarka2021ClassificationOM}, a mental illnesses dataset extracted from the Reddit social media platform. The task is to classify the post into one of the six following mental illnesses Bipolar, ADHD, anxiety, depression, PTSD, and None. The dataset consists of 16703 social media posts from Reddit, which include the title, the post, and the illness. The dataset is split into three parts, training - 13727, validation - 1488, and test - 1488 posts. We used both titles and posts combined for our analysis.
\subsection{System Description and Ensembling}
\label{subsection:sde}
We have used various models for our experiments, including:
BERT \cite{Devlin2019BERTPO} is a transformer-based \cite{Vaswani2017AttentionIA} language model. RoBERTa \cite{Liu2019RoBERTaAR} a faster variation of BERT. DeBERTa \cite{He2021DeBERTaDB} variant of RoBERTa. AlBERT \cite{Lan2020ALBERTAL} is a lightweight Self-supervised variant of BERT. We performed ablations with many of the present well-known language models — including Electra \cite{Clark2020ELECTRAPT}, and DistilBERT \cite{Sanh2019DistilBERTAD}.
We conducted extensive experimentation and observed some models to perform substantially better than others.
We shortlisted the models based on Accuracy and F1-Score on the valuation set for our proposed architecture. The shortlisted variants are BERT, RoBERTa, DeBERTa, and AlBERT.
\section{Results}
We evaluate the results from the baseline models and the proposed architecture over a fixed test set using Accuracy and F1 Score. We use three metrics to quantify the uncertainty - Expected Calibration Error (ECE), Maximum Calibration Error (MCE)\cite{NEURIPS2019_1c336b80}, and Brier Score\cite{rufibach2010use}. More Information about them is given in Appendix \ref{section:appA}. Table \ref{tab:comb} shows that all the baseline models and proposed model architecture have quite similar accuracy and F1 score. However, it is observed from the uncertainty metrics that the baseline models are not well-calibrated and reliable enough. However, using our proposed architecture, we observe an instant drop in Uncertainty Metrics, thus producing quite reliable and well-calibrated models.
\section{Conclusion and Future Works}
Through the proposed model architecture, we aim to produce more reliable model predictions by reducing model calibration error and improving reliability. We ensemble predictions from multiple models and generalize to OOD samples due to TTAug Section \ref{subsection:tta}. We believe there is immense potential in the proposed architecture to be incorporated into various tasks using Bayesian methods. We hope our work helps the community in detecting and helping the people suffering from mental illness and inspires further research on uncertainty-aware modeling and reliable models for healthcare diagnosis.

\bibliography{References}

\begin{thebibliography}{}

\bibitem[\protect\citename{Ameer \bgroup \em et al.\egroup ,
  }2022]{Ameer2022MentalIC}
Iqra Ameer, Muhammad Arif, Grigori Sidorov, Helena G{\'o}mez-Adorno, and
  Alexander~F. Gelbukh.
\newblock Mental illness classification on social media texts using deep
  learning and transfer learning.
\newblock {\em ArXiv}, abs/2207.01012, 2022.

\bibitem[\protect\citename{Amjad \bgroup \em et al.\egroup ,
  }2020]{Amjad2020BendTT}
Maaz Amjad, Grigori Sidorov, Alisa Zhila, Helena G{\'o}mez-Adorno, Ilia
  Voronkov, and Alexander Gelbukh.
\newblock "bend the truth": Benchmark dataset for fake news detection in urdu
  language and its evaluation.
\newblock {\em J. Intell. Fuzzy Syst.}, 39:2457--2469, 2020.

\bibitem[\protect\citename{Amjad \bgroup \em et al.\egroup ,
  }2021]{Amjad2021ThreateningLD}
Maaz Amjad, Noman Ashraf, Alisa Zhila, Grigori Sidorov, Arkaitz Zubiaga, and
  Alexander~F. Gelbukh.
\newblock Threatening language detection and target identification in urdu
  tweets.
\newblock {\em IEEE Access}, 9:128302--128313, 2021.

\bibitem[\protect\citename{Andreas, }2020]{Andreas2020GoodEnoughCD}
Jacob Andreas.
\newblock Good-enough compositional data augmentation.
\newblock In {\em ACL}, 2020.

\bibitem[\protect\citename{Ayhan \bgroup \em et al.\egroup ,
  }2020]{Ayhan2020ExpertvalidatedEO}
Murat~Seçkin Ayhan, Laura Kuehlewein, Gulnar Aliyeva, Werner Inhoffen, Focke
  Ziemssen, and Philipp Berens.
\newblock Expert-validated estimation of diagnostic uncertainty for deep neural
  networks in diabetic retinopathy detection.
\newblock {\em Medical image analysis}, 64:101724, 2020.

\bibitem[\protect\citename{Benton \bgroup \em et al.\egroup ,
  }2017]{Benton2017MultiTaskLF}
Adrian Benton, Margaret Mitchell, and Dirk Hovy.
\newblock Multi-task learning for mental health using social media text.
\newblock {\em ArXiv}, abs/1712.03538, 2017.

\bibitem[\protect\citename{Blum \bgroup \em et al.\egroup ,
  }2015]{Blum2015VariationalDA}
Avrim Blum, Nika Haghtalab, and Ariel~D. Procaccia.
\newblock Variational dropout and the local reparameterization trick.
\newblock In {\em NIPS}, 2015.

\bibitem[\protect\citename{Blundell \bgroup \em et al.\egroup ,
  }2015]{pmlr-v37-blundell15}
Charles Blundell, Julien Cornebise, Koray Kavukcuoglu, and Daan Wierstra.
\newblock Weight uncertainty in neural network.
\newblock In Francis Bach and David Blei, editors, {\em Proceedings of the 32nd
  International Conference on Machine Learning}, volume~37 of {\em Proceedings
  of Machine Learning Research}, pages 1613--1622, Lille, France, 07--09 Jul
  2015. PMLR.

\bibitem[\protect\citename{Boyd, }2018]{Boyd2018UsingWE}
Adriane Boyd.
\newblock Using wikipedia edits in low resource grammatical error correction.
\newblock In {\em NUT@EMNLP}, 2018.

\bibitem[\protect\citename{Chang \bgroup \em et al.\egroup ,
  }2017]{chang2017active}
Haw-Shiuan Chang, Erik Learned-Miller, and Andrew McCallum.
\newblock Active bias: Training more accurate neural networks by emphasizing
  high variance samples.
\newblock {\em Advances in Neural Information Processing Systems}, 30, 2017.

\bibitem[\protect\citename{Clark \bgroup \em et al.\egroup ,
  }2020]{Clark2020ELECTRAPT}
Kevin Clark, Minh-Thang Luong, Quoc~V. Le, and Christopher~D. Manning.
\newblock Electra: Pre-training text encoders as discriminators rather than
  generators.
\newblock {\em ArXiv}, abs/2003.10555, 2020.

\bibitem[\protect\citename{Cohen \bgroup \em et al.\egroup ,
  }2019]{Cohen2019CertifiedAR}
Jeremy~M. Cohen, Elan Rosenfeld, and J.~Zico Kolter.
\newblock Certified adversarial robustness via randomized smoothing.
\newblock In {\em ICML}, 2019.

\bibitem[\protect\citename{Coppersmith \bgroup \em et al.\egroup ,
  }2015]{coppersmith-etal-2015-clpsych}
Glen Coppersmith, Mark Dredze, Craig Harman, Kristy Hollingshead, and Margaret
  Mitchell.
\newblock {CLP}sych 2015 shared task: Depression and {PTSD} on {T}witter.
\newblock In {\em Proceedings of the 2nd Workshop on Computational Linguistics
  and Clinical Psychology: From Linguistic Signal to Clinical Reality}, pages
  31--39, Denver, Colorado, June 5 2015. Association for Computational
  Linguistics.

\bibitem[\protect\citename{Dai and Adel, }2020]{Dai2020AnAO}
Xiang Dai and Heike Adel.
\newblock An analysis of simple data augmentation for named entity recognition.
\newblock {\em ArXiv}, abs/2010.11683, 2020.

\bibitem[\protect\citename{Devlin \bgroup \em et al.\egroup ,
  }2019]{Devlin2019BERTPO}
Jacob Devlin, Ming-Wei Chang, Kenton Lee, and Kristina Toutanova.
\newblock Bert: Pre-training of deep bidirectional transformers for language
  understanding.
\newblock In {\em NAACL}, 2019.

\bibitem[\protect\citename{Ding \bgroup \em et al.\egroup ,
  }2020]{Ding2020DAGADA}
Bosheng Ding, Linlin Liu, Lidong Bing, Canasai Kruengkrai, Thien~Hai Nguyen,
  Shafiq~R. Joty, Luo Si, and Chunyan Miao.
\newblock Daga: Data augmentation with a generation approach for low-resource
  tagging tasks.
\newblock In {\em EMNLP}, 2020.

\bibitem[\protect\citename{Dusenberry \bgroup \em et al.\egroup ,
  }2020]{Dusenberry2020AnalyzingTR}
Michael~W. Dusenberry, Dustin Tran, E.~Choi, Jonas Kemp, Jeremy Nixon, Ghassen
  Jerfel, Katherine~A. Heller, and Andrew~M. Dai.
\newblock Analyzing the role of model uncertainty for electronic health
  records.
\newblock {\em Proceedings of the ACM Conference on Health, Inference, and
  Learning}, 2020.

\bibitem[\protect\citename{Gal and Ghahramani, }2016]{Gal2016DropoutAA}
Yarin Gal and Zoubin Ghahramani.
\newblock Dropout as a bayesian approximation: Representing model uncertainty
  in deep learning.
\newblock {\em ArXiv}, abs/1506.02142, 2016.

\bibitem[\protect\citename{Gkotsis \bgroup \em et al.\egroup ,
  }2017]{Gkotsis2017CharacterisationOM}
George Gkotsis, Anika Oellrich, Sumithra Velupillai, Maria Liakata, Tim J.~P.
  Hubbard, Richard J.~B. Dobson, and Rina Dutta.
\newblock Characterisation of mental health conditions in social media using
  informed deep learning.
\newblock {\em Scientific Reports}, 7, 2017.

\bibitem[\protect\citename{Graves, }2011]{Graves2011PracticalVI}
Alex Graves.
\newblock Practical variational inference for neural networks.
\newblock In {\em NIPS}, 2011.

\bibitem[\protect\citename{Guo \bgroup \em et al.\egroup , }2017]{Guo2017OnCO}
Chuan Guo, Geoff Pleiss, Yu~Sun, and Kilian~Q. Weinberger.
\newblock On calibration of modern neural networks.
\newblock {\em ArXiv}, abs/1706.04599, 2017.

\bibitem[\protect\citename{Hamilton, }1967]{Hamilton1967DevelopmentOA}
Max Hamilton.
\newblock Development of a rating scale for primary depressive illness.
\newblock {\em The British journal of social and clinical psychology}, 6
  4:278--96, 1967.

\bibitem[\protect\citename{He \bgroup \em et al.\egroup ,
  }2021]{He2021DeBERTaDB}
Pengcheng He, Xiaodong Liu, Jianfeng Gao, and Weizhu Chen.
\newblock Deberta: Decoding-enhanced bert with disentangled attention.
\newblock {\em ArXiv}, abs/2006.03654, 2021.

\bibitem[\protect\citename{Hendrycks \bgroup \em et al.\egroup ,
  }2020]{Hendrycks2020PretrainedTI}
Dan Hendrycks, Xiaoyuan Liu, Eric Wallace, Adam Dziedzic, Rishabh Krishnan, and
  Dawn~Xiaodong Song.
\newblock Pretrained transformers improve out-of-distribution robustness.
\newblock In {\em ACL}, 2020.

\bibitem[\protect\citename{Hern{\'a}ndez-Lobato and Adams,
  }2015]{HernndezLobato2015ProbabilisticBF}
Jos{\'e}~Miguel Hern{\'a}ndez-Lobato and Ryan~P. Adams.
\newblock Probabilistic backpropagation for scalable learning of bayesian
  neural networks.
\newblock In {\em ICML}, 2015.

\bibitem[\protect\citename{Husseini~Orabi \bgroup \em et al.\egroup ,
  }2018]{husseini-orabi-etal-2018-deep}
Ahmed Husseini~Orabi, Prasadith Buddhitha, Mahmoud Husseini~Orabi, and Diana
  Inkpen.
\newblock Deep learning for depression detection of {T}witter users.
\newblock In {\em Proceedings of the Fifth Workshop on Computational
  Linguistics and Clinical Psychology: From Keyboard to Clinic}, pages 88--97,
  New Orleans, LA, June 2018. Association for Computational Linguistics.

\bibitem[\protect\citename{Jaskari \bgroup \em et al.\egroup ,
  }2022]{Jaskari2022UncertaintyAwareDL}
Joel Jaskari, Jaakko Sahlsten, Theodoros Damoulas, Jeremias Knoblauch, Simo
  S{\"a}rkk{\"a}, L.~K{\"a}rkk{\"a}inen, Kustaa Hietala, and Kimmo~K. Kaski.
\newblock Uncertainty-aware deep learning methods for robust diabetic
  retinopathy classification.
\newblock {\em IEEE Access}, 10:76669--76681, 2022.

\bibitem[\protect\citename{Jia and Liang, }2016]{Jia2016DataRF}
Robin Jia and Percy Liang.
\newblock Data recombination for neural semantic parsing.
\newblock {\em ArXiv}, abs/1606.03622, 2016.

\bibitem[\protect\citename{Kendall \bgroup \em et al.\egroup ,
  }2018]{Kendall_2018_CVPR}
Alex Kendall, Yarin Gal, and Roberto Cipolla.
\newblock Multi-task learning using uncertainty to weigh losses for scene
  geometry and semantics.
\newblock In {\em Proceedings of the IEEE Conference on Computer Vision and
  Pattern Recognition (CVPR)}, June 2018.

\bibitem[\protect\citename{Kim \bgroup \em et al.\egroup , }2020]{Kim2020ADL}
Jina Kim, Jieon Lee, Eunil Park, and Jinyoung Han.
\newblock A deep learning model for detecting mental illness from user content
  on social media.
\newblock {\em Scientific Reports}, 10, 2020.

\bibitem[\protect\citename{Kingma and Ba, }2015]{Kingma2015AdamAM}
Diederik~P. Kingma and Jimmy Ba.
\newblock Adam: A method for stochastic optimization.
\newblock {\em CoRR}, abs/1412.6980, 2015.

\bibitem[\protect\citename{Krizhevsky \bgroup \em et al.\egroup ,
  }2012]{Krizhevsky2012ImageNetCW}
Alex Krizhevsky, Ilya Sutskever, and Geoffrey~E. Hinton.
\newblock Imagenet classification with deep convolutional neural networks.
\newblock {\em Communications of the ACM}, 60:84 -- 90, 2012.

\bibitem[\protect\citename{Lakara \bgroup \em et al.\egroup ,
  }2021]{Lakara2021EvaluatingPU}
Kumud Lakara, Akshat Bhandari, Pratinav Seth, and Ujjwal Verma.
\newblock Evaluating predictive uncertainty and robustness to distributional
  shift using real world data.
\newblock {\em ArXiv}, abs/2111.04665, 2021.

\bibitem[\protect\citename{Lakshminarayanan \bgroup \em et al.\egroup ,
  }2017]{Lakshminarayanan2017SimpleAS}
Balaji Lakshminarayanan, Alexander Pritzel, and Charles Blundell.
\newblock Simple and scalable predictive uncertainty estimation using deep
  ensembles.
\newblock In {\em NIPS}, 2017.

\bibitem[\protect\citename{Lan \bgroup \em et al.\egroup ,
  }2020]{Lan2020ALBERTAL}
Zhenzhong Lan, Mingda Chen, Sebastian Goodman, Kevin Gimpel, Piyush Sharma, and
  Radu Soricut.
\newblock Albert: A lite bert for self-supervised learning of language
  representations.
\newblock {\em ArXiv}, abs/1909.11942, 2020.

\bibitem[\protect\citename{Lee \bgroup \em et al.\egroup ,
  }2018]{Lee2018DeepNN}
Jaehoon Lee, Yasaman Bahri, Roman Novak, Samuel~S. Schoenholz, Jeffrey
  Pennington, and Jascha~Narain Sohl-Dickstein.
\newblock Deep neural networks as gaussian processes.
\newblock {\em ArXiv}, abs/1711.00165, 2018.

\bibitem[\protect\citename{Liu \bgroup \em et al.\egroup ,
  }2019]{Liu2019RoBERTaAR}
Yinhan Liu, Myle Ott, Naman Goyal, Jingfei Du, Mandar Joshi, Danqi Chen, Omer
  Levy, Mike Lewis, Luke Zettlemoyer, and Veselin Stoyanov.
\newblock Roberta: A robustly optimized bert pretraining approach.
\newblock {\em ArXiv}, abs/1907.11692, 2019.

\bibitem[\protect\citename{Longpre \bgroup \em et al.\egroup ,
  }2019]{Longpre2019AnEO}
S.~Longpre, Yi~Lu, Zhucheng Tu, and Christopher DuBois.
\newblock An exploration of data augmentation and sampling techniques for
  domain-agnostic question answering.
\newblock In {\em EMNLP}, 2019.

\bibitem[\protect\citename{Lu \bgroup \em et al.\egroup ,
  }2022]{Lu2022ImprovedTC}
Helen~Shiyang Lu, Divya Shanmugam, Harini Suresh, and John~V. Guttag.
\newblock Improved text classification via test-time augmentation.
\newblock {\em ArXiv}, abs/2206.13607, 2022.

\bibitem[\protect\citename{Ma, }2019]{ma2019nlpaug}
Edward Ma.
\newblock Nlp augmentation.
\newblock https://github.com/makcedward/nlpaug, 2019.

\bibitem[\protect\citename{Matsunaga \bgroup \em et al.\egroup ,
  }2017]{Matsunaga2017ImageCO}
Kazuhisa Matsunaga, Akira Hamada, Akane Minagawa, and Hiroshi Koga.
\newblock Image classification of melanoma, nevus and seborrheic keratosis by
  deep neural network ensemble.
\newblock {\em ArXiv}, abs/1703.03108, 2017.

\bibitem[\protect\citename{Murarka \bgroup \em et al.\egroup ,
  }2021]{Murarka2021ClassificationOM}
Ankit Murarka, Balaji Radhakrishnan, and Sushma Ravichandran.
\newblock Classification of mental illnesses on social media using roberta.
\newblock {\em ArXiv}, abs/2011.11226, 2021.

\bibitem[\protect\citename{Osband, }2016]{Osband2016RiskVU}
Ian Osband.
\newblock Risk versus uncertainty in deep learning: Bayes, bootstrap and the
  dangers of dropout.
\newblock {\em Workshop on Bayesian Deep Learning, NIPS}, 2016.

\bibitem[\protect\citename{Ovadia \bgroup \em et al.\egroup ,
  }2019]{Ovadia2019CanYT}
Yaniv Ovadia, Emily Fertig, J.~Ren, Zachary Nado, D.~Sculley, Sebastian
  Nowozin, Joshua~V. Dillon, Balaji Lakshminarayanan, and Jasper Snoek.
\newblock Can you trust your model's uncertainty? evaluating predictive
  uncertainty under dataset shift.
\newblock {\em ArXiv}, abs/1906.02530, 2019.

\bibitem[\protect\citename{Parida and Motl{\'i}cek,
  }2019]{Parida2019AbstractTS}
Shantipriya Parida and Petr Motl{\'i}cek.
\newblock Abstract text summarization: A low resource challenge.
\newblock In {\em EMNLP}, 2019.

\bibitem[\protect\citename{Pearce \bgroup \em et al.\egroup ,
  }2020]{Pearce2020UncertaintyIN}
Tim Pearce, Felix Leibfried, Alexandra Brintrup, Mohamed Zaki, and A.~D. Neely.
\newblock Uncertainty in neural networks: Approximately bayesian ensembling.
\newblock In {\em AISTATS}, 2020.

\bibitem[\protect\citename{Prakash \bgroup \em et al.\egroup ,
  }2018]{Prakash2018DeflectingAA}
Aaditya Prakash, Nick Moran, Solomon Garber, Antonella DiLillo, and James~A.
  Storer.
\newblock Deflecting adversarial attacks with pixel deflection.
\newblock {\em 2018 IEEE/CVF Conference on Computer Vision and Pattern
  Recognition}, pages 8571--8580, 2018.

\bibitem[\protect\citename{Rufibach, }2010]{rufibach2010use}
Kaspar Rufibach.
\newblock Use of brier score to assess binary predictions.
\newblock {\em Journal of clinical epidemiology}, 63(8):938--939, 2010.

\bibitem[\protect\citename{Sahin and Steedman, }2018]{Sahin2018DataAV}
G{\"o}zde~G{\"u}l Sahin and Mark Steedman.
\newblock Data augmentation via dependency tree morphing for low-resource
  languages.
\newblock In {\em EMNLP}, 2018.

\bibitem[\protect\citename{Sanh \bgroup \em et al.\egroup ,
  }2019]{Sanh2019DistilBERTAD}
Victor Sanh, Lysandre Debut, Julien Chaumond, and Thomas Wolf.
\newblock Distilbert, a distilled version of bert: smaller, faster, cheaper and
  lighter.
\newblock {\em ArXiv}, abs/1910.01108, 2019.

\bibitem[\protect\citename{Sarawgi \bgroup \em et al.\egroup ,
  }2020]{Sarawgi2020WhyHA}
Utkarsh Sarawgi, Wazeer Zulfikar, Rishab Khincha, and Pattie Maes.
\newblock Why have a unified predictive uncertainty? disentangling it using
  deep split ensembles.
\newblock {\em ArXiv}, abs/2009.12406, 2020.

\bibitem[\protect\citename{Sarawgi \bgroup \em et al.\egroup ,
  }2021]{sarawgi2021uncertainty}
Utkarsh Sarawgi, Rishab Khincha, Wazeer Zulfikar, Satrajit Ghosh, and Pattie
  Maes.
\newblock Uncertainty-aware boosted ensembling in multi-modal settings.
\newblock In {\em 2021 International Joint Conference on Neural Networks
  (IJCNN)}, pages 1--9. IEEE, 2021.

\bibitem[\protect\citename{Sayers, }2001]{Sayers2001TheWH}
Janet~V. Sayers.
\newblock The world health report 2001 - mental health: new understanding, new
  hope.
\newblock {\em Bulletin of The World Health Organization}, 79:1085--1085, 2001.

\bibitem[\protect\citename{Sennrich \bgroup \em et al.\egroup ,
  }2016]{Sennrich2016ImprovingNM}
Rico Sennrich, Barry Haddow, and Alexandra Birch.
\newblock Improving neural machine translation models with monolingual data.
\newblock {\em ArXiv}, abs/1511.06709, 2016.

\bibitem[\protect\citename{Shanmugam \bgroup \em et al.\egroup ,
  }2021]{Shanmugam2021BetterAI}
Divya Shanmugam, Davis~W. Blalock, Guha Balakrishnan, and John~V. Guttag.
\newblock Better aggregation in test-time augmentation.
\newblock {\em 2021 IEEE/CVF International Conference on Computer Vision
  (ICCV)}, pages 1194--1203, 2021.

\bibitem[\protect\citename{Singh \bgroup \em et al.\egroup ,
  }2019]{Singh2019XLDACD}
Jasdeep Singh, Bryan McCann, Nitish~Shirish Keskar, Caiming Xiong, and Richard
  Socher.
\newblock Xlda: Cross-lingual data augmentation for natural language inference
  and question answering.
\newblock {\em ArXiv}, abs/1905.11471, 2019.

\bibitem[\protect\citename{Smith and Gal, }2018]{Smith2018UnderstandingMO}
Lewis Smith and Yarin Gal.
\newblock Understanding measures of uncertainty for adversarial example
  detection.
\newblock {\em ArXiv}, abs/1803.08533, 2018.

\bibitem[\protect\citename{Song \bgroup \em et al.\egroup ,
  }2018]{Song2018PixelDefendLG}
Yang Song, Taesup Kim, Sebastian Nowozin, Stefano Ermon, and Nate Kushman.
\newblock Pixeldefend: Leveraging generative models to understand and defend
  against adversarial examples.
\newblock {\em ArXiv}, abs/1710.10766, 2018.

\bibitem[\protect\citename{Szegedy \bgroup \em et al.\egroup ,
  }2015]{Szegedy2015GoingDW}
Christian Szegedy, Wei Liu, Yangqing Jia, Pierre Sermanet, Scott~E. Reed,
  Dragomir Anguelov, D.~Erhan, Vincent Vanhoucke, and Andrew Rabinovich.
\newblock Going deeper with convolutions.
\newblock {\em 2015 IEEE Conference on Computer Vision and Pattern Recognition
  (CVPR)}, pages 1--9, 2015.

\bibitem[\protect\citename{Vaswani \bgroup \em et al.\egroup ,
  }2017]{Vaswani2017AttentionIA}
Ashish Vaswani, Noam~M. Shazeer, Niki Parmar, Jakob Uszkoreit, Llion Jones,
  Aidan~N. Gomez, Lukasz Kaiser, and Illia Polosukhin.
\newblock Attention is all you need.
\newblock In {\em NIPS}, 2017.

\bibitem[\protect\citename{Wang \bgroup \em et al.\egroup ,
  }2019]{Wang2019AleatoricUE}
Guotai Wang, Wenqi Li, Michael Aertsen, Jan~A. Deprest, S{\'e}bastien Ourselin,
  and Tom Kamiel~Magda Vercauteren.
\newblock Aleatoric uncertainty estimation with test-time augmentation for
  medical image segmentation with convolutional neural networks.
\newblock {\em Neurocomputing}, 335:34 -- 45, 2019.

\bibitem[\protect\citename{Wang \bgroup \em et al.\egroup ,
  }2022]{Wang2022ControllableDS}
Chencheng Wang, Liner Yang, Yuxiang Chen, Yong ping Du, and Erhong Yang.
\newblock Controllable data synthesis method for grammatical error correction.
\newblock {\em Frontiers Comput. Sci.}, 16:164318, 2022.

\bibitem[\protect\citename{Widmann \bgroup \em et al.\egroup ,
  }2019]{NEURIPS2019_1c336b80}
David Widmann, Fredrik Lindsten, and Dave Zachariah.
\newblock Calibration tests in multi-class classification: A unifying
  framework.
\newblock In H.~Wallach, H.~Larochelle, A.~Beygelzimer, F.~d\textquotesingle
  Alch\'{e}-Buc, E.~Fox, and R.~Garnett, editors, {\em Advances in Neural
  Information Processing Systems}, volume~32. Curran Associates, Inc., 2019.

\bibitem[\protect\citename{Wolf \bgroup \em et al.\egroup ,
  }2020]{wolf-etal-2020-transformers}
Thomas Wolf, Lysandre Debut, Victor Sanh, Julien Chaumond, Clement Delangue,
  Anthony Moi, Pierric Cistac, Tim Rault, Rémi Louf, Morgan Funtowicz, Joe
  Davison, Sam Shleifer, Patrick von Platen, Clara Ma, Yacine Jernite, Julien
  Plu, Canwen Xu, Teven~Le Scao, Sylvain Gugger, Mariama Drame, Quentin Lhoest,
  and Alexander~M. Rush.
\newblock Transformers: State-of-the-art natural language processing.
\newblock In {\em Proceedings of the 2020 Conference on Empirical Methods in
  Natural Language Processing: System Demonstrations}, pages 38--45, Online,
  October 2020. Association for Computational Linguistics.

\bibitem[\protect\citename{Wu \bgroup \em et al.\egroup ,
  }2019]{Wu2019DeterministicVI}
Anqi Wu, Sebastian Nowozin, Edward Meeds, Richard~E. Turner, Jos{\'e}~Miguel
  Hern{\'a}ndez-Lobato, and Alexander~L. Gaunt.
\newblock Deterministic variational inference for robust bayesian neural
  networks.
\newblock In {\em ICLR}, 2019.

\bibitem[\protect\citename{Xia \bgroup \em et al.\egroup ,
  }2019]{Xia2019GeneralizedDA}
M.~Xia, X.~Kong, Antonios Anastasopoulos, and Graham Neubig.
\newblock Generalized data augmentation for low-resource translation.
\newblock {\em ArXiv}, abs/1906.03785, 2019.

\bibitem[\protect\citename{Yang and Fevens, }2021]{Yang2021UncertaintyQA}
Sidi Yang and T.~Fevens.
\newblock Uncertainty quantification and estimation in medical image
  classification.
\newblock In {\em ICANN}, 2021.

\bibitem[\protect\citename{Yang \bgroup \em et al.\egroup ,
  }2019]{Yang2019DataAF}
Wei Yang, Yuqing Xie, Luchen Tan, Kun Xiong, Ming Li, and Jimmy~J. Lin.
\newblock Data augmentation for bert fine-tuning in open-domain question
  answering.
\newblock {\em ArXiv}, abs/1904.06652, 2019.

\bibitem[\protect\citename{Yu \bgroup \em et al.\egroup ,
  }2021]{Yu2021GraPPaGP}
Tao Yu, Chien-Sheng Wu, Xi~Victoria Lin, Bailin Wang, Yi~Chern Tan, Xinyi Yang,
  Dragomir Radev, Richard Socher, and Caiming Xiong.
\newblock Grappa: Grammar-augmented pre-training for table semantic parsing.
\newblock {\em ArXiv}, abs/2009.13845, 2021.

\bibitem[\protect\citename{Zhang \bgroup \em et al.\egroup ,
  }2019]{Zhang2019SequencetosequencePW}
Yi~Zhang, Tao Ge, Furu Wei, Ming Zhou, and Xu~Sun.
\newblock Sequence-to-sequence pre-training with data augmentation for sentence
  rewriting.
\newblock {\em ArXiv}, abs/1909.06002, 2019.

\bibitem[\protect\citename{Zhu \bgroup \em et al.\egroup ,
  }2022]{Zhu2022TransformingWI}
Haichao Zhu, Li~Dong, Furu Wei, Bing Qin, and Ting Liu.
\newblock Transforming wikipedia into augmented data for query-focused
  summarization.
\newblock {\em IEEE/ACM Transactions on Audio, Speech, and Language
  Processing}, 30:2357--2367, 2022.

\bibitem[\protect\citename{Zirikly \bgroup \em et al.\egroup ,
  }2019]{Zirikly2019CLPsych2S}
Ayah Zirikly, Philip Resnik, {\"O}zlem Uzuner, and Kristy Hollingshead.
\newblock Clpsych 2019 shared task: Predicting the degree of suicide risk in
  reddit posts.
\newblock {\em Proceedings of the Sixth Workshop on Computational Linguistics
  and Clinical Psychology}, 2019.

\end{thebibliography}
\appendix
\section{Appendix} 
\label{apd:first}
\label{section:appA}
\subsection{Evaluation and Uncertainty Metrics}
The model performance is evaluated using Accuracy and F1 Score.We use three metrics to quantify the uncertainty - Expected Calibration Error (ECE), Maximum Calibration Error (MCE) (\cite{NEURIPS2019_1c336b80}), and Brier Score (\cite{rufibach2010use}).

\subsubsection{Expected Calibration Error}
The Expected Calibration Error (ECE) is a weighted average over the absolute confidence difference of the predictions of a model. It is defined as 
\begin{equation}
\label{sec:ece}
 ECE =  \sum_{i \in \mathcal{B_m}} \frac{|B_m|}{n} |acc(B_m)-conf(B_m)|
\end{equation}
where
\begin{equation}
\label{sec:ece1}
  acc(B_m) =  \frac{1}{|B_m|} \sum_{i \in B_m}  1(y_{i}=y_{t})
\end{equation}
\begin{equation}
\label{sec:ece2}
 conf(B_m) =  \frac{1}{|B_m|} \sum_{i \in B_m}  p_{i}
\end{equation}
where $conf(B_m)$ is just the average confidence/probability of predictions in that bin, and $acc(B_m)$ is the fraction of the correctly classified examples Bm.
\subsubsection{Maximum Calibration Error}
The Maximum Calibration Error (MCE) (\cite{NEURIPS2019_1c336b80}) focuses more on high-risk applications where the maximum confidence difference is more important than the average.
It is then defined as:
\begin{equation}
\label{sec:mce}
  MCE =  max_{m} |acc(B_m)-conf(B_m)|
\end{equation}
\subsubsection{Brier Score}

The Brier Score (\cite{rufibach2010use}) is a strictly proper score function or scoring rule that measures the accuracy of probabilistic predictions.
\begin{equation}
\label{sec:bs}
  Brier Score = \frac{1}{n} \sum_{t=1}^{n} (f_{t} - o_{t})^2 
\end{equation}
where f{t} is the probability that was forecast, o{t} is the actual outcome of the event at instance t (0 if it does not happen and one if it does happen), and N is the number of forecasting instances.

\newpage
\end{document}